\documentclass{article}
\usepackage{graphicx} %
\usepackage{amsmath}
\usepackage{amssymb}
\usepackage{amsfonts}
\usepackage{hyperref}
\usepackage{comment}
\usepackage{xcolor}
\usepackage{geometry}
\usepackage{listings}
\usepackage{float}
\usepackage{authblk}

\usepackage{multicol}  %

\usepackage[numbers]{natbib}

\begin{document}

\begin{center}
    \huge {Prompt Baking}
\end{center}

\vspace{3em} %

\begin{minipage}[t]{0.33\textwidth}
\centering
\textbf{Aman Bhargava$^{*\dagger}$} \\
{Caltech} \\
\texttt{abhargav@caltech.edu} \\
\end{minipage}%
\begin{minipage}[t]{0.33\textwidth}
\centering
\textbf{Cameron Witkowski$^{*\ddagger}$} \\
{University of Toronto} \\
\texttt{cameron.witkowski@\\mail.utoronto.ca} \\
\end{minipage}%
\begin{minipage}[t]{0.33\textwidth}
\centering
\textbf{Alexander Detkov$^{*\dagger}$} \\
{Caltech} \\
\texttt{adetkov@caltech.edu} \\
\end{minipage}

\vspace{2em} %

\begin{minipage}[t]{0.99\textwidth}
\centering
\textbf{Matt Thomson$^\dagger$} \\
{Caltech} \\
\texttt{matt@caltech.edu} \\
\end{minipage}

\vspace{1em} %

\renewcommand{\thefootnote}{\fnsymbol{footnote}} %
\footnotetext[1]{Equal contribution.}
\footnotetext[2]{Computation \& Neural Systems, California Institute of Technology.}
\footnotetext[3]{Department of Electrical and Computer Engineering, University of Toronto.}
\renewcommand{\thefootnote}{\arabic{footnote}} %

\section*{Abstract}

Two primary ways to change LLM behavior are prompting and weight updates (e.g. fine-tuning).
Prompting LLMs is simple and effective, specifying the desired changes explicitly in natural language, whereas weight updates provide more expressive and permanent behavior changes, specified implicitly via training on large datasets.
We present a technique for ``baking'' prompts into the weights of an LLM. 
\textsc{Prompt Baking} converts a prompt $\mathbf u$ and initial weights $\theta$ to a new set of weights $\theta_{\mathbf u}$ such that the LLM with weights $\theta_{\mathbf u}$ behaves like the LLM with weights $\theta$ and prompt $\mathbf u$.
Mathematically, we minimize the KL divergence between ${P}_\theta(\cdot | \mathbf u )$ and ${P}_{\theta_{\mathbf u}}(\cdot)$, where ${P}$ is the LLM's probability distribution over token sequences.
Across all our experiments, we find prompts can be readily baked into weight updates, often in as little as 5 minutes.
Baking chain-of-thought prompts improves zero-shot performance on \textsc{GSM8K, ASDiv, MBPP, ARC-Easy, ARC-Challenge}, and \textsc{CommonsenseQA} benchmarks.
Baking news headlines directly updates an LLM's knowledge. 
And baking instructions \& personas alleviates ``prompt forgetting'' over long sequences, as measured on a \textsc{Persona Drift} benchmark.
Furthermore, stopping baking early creates ``half-baked'' models, allowing for continuous scaling of prompt strength. 
Baked models retain their sensitivity to further prompting and baking, including \textbf{re-prompting} with the prompt already baked in -- thus amplifying the prompt's strength.
Surprisingly, the re-prompted models yield further performance gains in instruction following, as well as math reasoning and coding benchmarks (\textsc{GSM8K, ASDiv,} and \textsc{MBPP}). 
Taking re-prompting and re-baking to the limit yields a form of iterative self-improvement we call \textsc{Prompt Pursuit}, and preliminary results on instruction following exhibit dramatic performance gains with this technique.
Finally, we discuss implications for AI safety, continuous model updating, improving LLM recency, enhancing real-time learning capabilities in LLM-based agents, and methods for generating more stable AI personas.

\vspace{2em}

\section{Introduction}

Scale endows language models with remarkable generality and adaptability through prompt-based model programming (e.g., zero-shot capabilities).
However, true adaptability demands new learning methods that permanently incorporate new knowledge into the LLM without labor-intensive curated datasets.
Prompting \textbf{explicitly} communicates new information to an LLM in natural language, relying on the ``zero-shot'' capabilities of the underlying model \cite{brown2020languagemodelsfewshotlearners}. 
Information learned via weight updates (e.g., fine-tuning) is \textbf{implicitly} communicated through the statistics of a data distribution \cite{goodfellow_dl}. 
While weight updates are maximally expressive and permanent, it is challenging to reliably communicate new knowledge via weight updates as it is unclear what information implicit in the dataset will be learned by the model \cite{raghavan2023engineeringflexiblemachinelearning, DBLP:journals/corr/abs-2107-10295, wang2024largelanguagemodelenhanced}. 
Moreover, the original model's robustness, abilities, and knowledge are not guaranteed to carry over to the fine-tuned version.
While prompting is extremely low-cost and often effective, models are unable to retain knowledge from prompts, and it remains challenging to ``navigate'' the discrete, and often highly chaotic, space of prompts to reliably and precisely control a language model \cite{bhargava2024whatsmagicwordcontrol}. Prompting also occupies valuable attentional space within the context of an LLM \cite{jiang2023llmlinguacompressingpromptsaccelerated}, and the influence of prompts over subsequent text generation has been shown to decay with long sequences\cite{li2024measuringcontrollinginstructioninstability}.\\

We introduce \textsc{Prompt Baking}: a technique to convert prompts into weight updates for an LLM. 
Prompt baking maps an initial set of LLM weights $\theta$ and prompt $\mathbf u$ to a new set of weights $\theta_{\mathbf u}$ such that the behavior of an LLM with weights $\theta_{\mathbf u}$ is similar to an LLM with weights $\theta$ prompted with $\mathbf u$. 
The specific optimization objective is to minimize the KL divergence $D_{KL}$ between the original prompted model (weights $\theta$, prompt $\mathbf u$) and the new ``baked'' model (weights $\theta_{\mathbf u}$, no prompt). 
Minimizing $D_{KL}$ essentially corresponds to matching the logits of the prompted original model $P_\theta(\cdot | \mathbf u)$ and the baked model $P_{\theta_{\mathbf u}}(\cdot)$. \\

\begin{figure}
    \centering
    \includegraphics[width=0.65\linewidth]{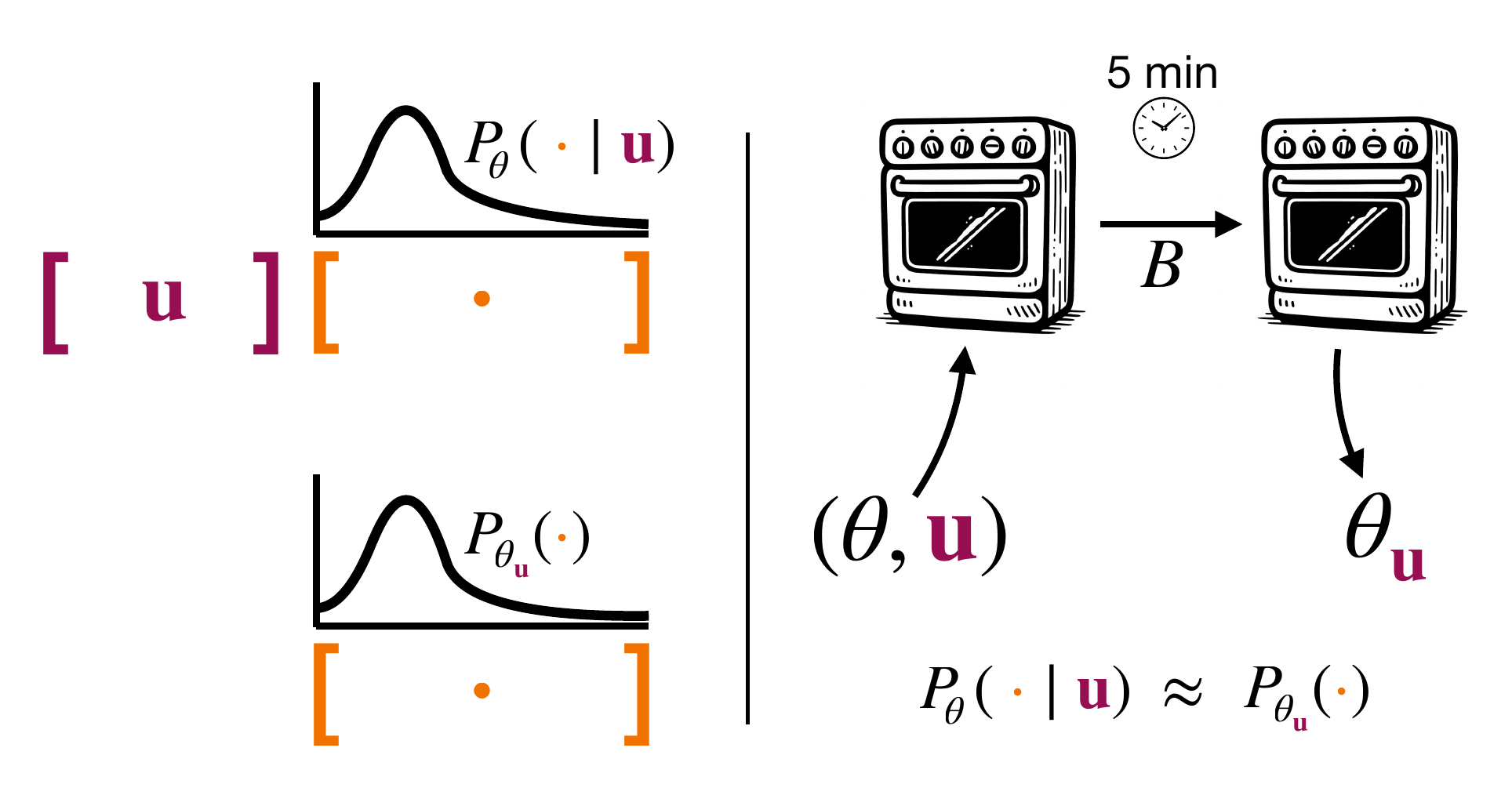}
    \caption{An illustration of \textsc{Prompt Baking}.}
    \label{fig:prompt_baking}
\end{figure}

\subsection{Contribution} 
\begin{itemize}
    \item \textbf{Prompt baking works: } We demonstrate \textsc{Prmopt Baking} on a number of tasks and academic benchmarks, validating that the baked model {$P_{\theta_{\mathbf u}}(\cdot)$ matches the behavior of the prompted baseline model $P_{\theta}(\cdot | \mathbf u)$ (Section~\ref{sec:r2})}. 
    We showcase how baking transforms ephemeral prompts into durable weight updates, ensuring consistent model behavior across multiple uses without repeated prompting. Moreover, prompts can often be baked as little as 5 minutes of wall clock time, showcasing the practicality and usability of our approach.
    
    \item \textbf{Scaling prompts: } Our technique allows continuous interpolation between the pre-baking and fully baked states, something we call ``half-baking''. 
    This capability addresses a long-standing challenge in prompt engineering: the difficulty of navigating the discrete and chaotic space of prompts, where subtle adjustments to prompt strength are often hard to achieve (Section~\ref{sec:scaling}).
    Moreover, by repeatedly applying the prompt baking operator, we can amplify a prompt's influence beyond its original strength, leading to further performance gains on a variety of tasks (Section~\ref{sec:pursuit}).

    \item \textbf{Baking in chain-of-thought examples on Academic Benchmarks: } We demonstrate that baking in few-shot chain-of-thought prompts attains nearly the same performance boost as few-shot prompting on \textsc{GSM8K, ASDiv, MBPP, ARC-Easy, ARC-Challenge}, and \textsc{CommonsenseQA} benchmarks. 
    We find that baked models achieve accuracy within 1.4\% of the corresponding few-shot prompted model on every benchmark.
    Shockingly, on some benchmarks, baking then prompting the baked model \textit{surpasses} the performance of prompting alone. Most notably, on GSM8K, we surpass Meta's published accuracy by 1.4\%, while using the same fewshot examples and experimental setup.
    
    \item \textbf{Composing prompts/knowledge baking: } Our technique successfully bakes multiple previously unseen news headlines sequentially, effectively baking in new distinct pieces of information (Section~\ref{sec:knowledge_baking}). We demonstrate this knowledge that baked knowledge is not merely memorized on a surface level by eliciting it with indirect queries post-baking.
    These results open the potential to incorporate more diverse knowledge than traditional single-prompt approaches, suggesting a path to overcome context window limitations.
    
    \item \textbf{Ameliorates prompt decay: } By integrating prompt information directly into model weights, prompt baking alleviates the decay of prompt influence over long sequences. 
    This results in consistent performance throughout extended dialogues, addressing a key limitation of traditional prompting methods (Section~\ref{sec:decay}).
\end{itemize}

\subsection{Related Work}

\paragraph{Model Distillation: } Our work builds upon the principles of model distillation \cite{xu2024surveyknowledgedistillationlarge}, a technique that minimizes the KL divergence between a teacher and a student model. 
Traditional distillation matches logits on an existing corpus, leveraging the informativeness of soft targets. 
Recent work, such as ``Distilling Reasoning Capabilities into Smaller Language Models" \cite{shridhar2023distillingreasoningcapabilitiessmaller}, extends this approach by using generated chain-of-thought trajectories. 
Our method, however, distinguishes itself in several key aspects: (1) we distill from the same model, effectively `self-distilling' prompted behavior; (2) we enable the baking of any prompt, not limited to reasoning prompts; and (3) our generated trajectories do not require answer checking or even the presence of answers, thus is entirely self-supervised. 
This approach connects to the concept of self-improvement while eliminating the need for explicit answer validation, offering a more flexible and generalizable framework for knowledge incorporation.

\paragraph{LLM Control Theory: } Our previous work \cite{bhargava2024whatsmagicwordcontrol} introduced a control theoretic framework for LLM prompting, demonstrating high controllability via short prompts. 
\textsc{Prompt Baking} extends this by projecting prompt-induced probability changes onto weight space, effectively ``moving" the model's default behavior within the previously defined reachable set. 
Crucially, our half-baking approach allows for fine-grained control by enabling continuous interpolation between unprompted and fully prompted states. 
As the model minimizes KL divergence during baking, it traverses a path in weight space connecting these two states. 
This conceptualization aligns with the notion of functionally invariant paths (FIPs) in weight space \cite{raghavan2023engineeringflexiblemachinelearning}. 
While FIPs are constructed to maintain functional performance while optimizing secondary objectives, our method implicitly defines a path that ``bakes in'' the prompted behavior to varying degrees. 
This connection between prompting, weight updates, and weight space paths offers a powerful new perspective and technique on LLM control.

\paragraph{Self-Improvement: } Our work relates to recent developments in LLM self-improvement, such as those described in \cite{huang2022largelanguagemodelsselfimprove, wang2023selfinstructaligninglanguagemodels}. 
These approaches train LLMs on self-generated trajectories or instruction-input-output triplets to enhance performance. 
However, prompt baking differs in several crucial aspects. 
Unlike self-improvement methods that use cross-entropy loss and focus on specific types of outputs (e.g., high-confidence rationale-augmented answers), we employ KL divergence and can bake in any type of prompt. 
Our approach doesn't require majority voting, filtering of trajectories, or elimination of duplicates. 
Furthermore, while Self-Instruct generates and trains on instruction-following examples, we generate trajectories directly from the prompt being baked in. 
This makes our method more generalizable, capable of incorporating a wide range of prompts beyond just instruction-following, while maintaining a simpler training process without output validation or filtering.

\paragraph{Knowledge Editing: } Our work relates to the emerging field of knowledge editing in LLMs \cite{zhang2024comprehensivestudyknowledgeediting}. 
Knowledge editing aims to modify specific information or behaviors in LLMs without full retraining, addressing issues like factual errors or outdated information. 
While knowledge editing methods often focus on altering particular facts or responses, our prompt baking approach offers a more general framework for incorporating diverse prompts into model weights. 
Moreover, our method preserves the model's sensitivity to further prompting, a feature not typically emphasized in knowledge editing approaches. 
While knowledge editing often struggles with portability (applying edits to related contexts), we find that knowledge editing via prompt baking yields models able to access the new knowledge in a robust and general way, even when asked about it indirectly. 

\paragraph{Continual Learning and Catastrophic Forgetting: } Our work addresses key challenges in continual learning for AI systems, particularly the problem of catastrophic forgetting (CF) in large language models (LLMs) \cite{wang2024comprehensivesurveycontinuallearning, zenke2017continual, luo2024empiricalstudycatastrophicforgetting}. 
CF occurs when models rapidly forget previously learned information while acquiring new knowledge, a phenomenon recently observed in LLMs during continual instruction tuning, with severity increasing with model size \cite{MCCLOSKEY1989109, FRENCH1999128, luo2024empiricalstudycatastrophicforgetting}. 
Unlike traditional continual learning methods that often focus on task-specific updates or require experience replay, prompt baking offers a novel approach to mitigate CF.
It allows for targeted updates to LLM behavior without extensive retraining or explicit task boundaries, addressing the stability-plasticity trade-off by enabling precise knowledge incorporation while maintaining overall model performance. 
Our method demonstrates the ability to sequentially incorporate multiple pieces of knowledge (e.g., news headlines) without significant forgetting, a capability that many continual learning techniques struggle with. Additionally, prompt baking mitigates prompt decay over long sequences, indirectly contributing to CF reduction. 
The scalability and composability of our technique offer promising avenues for developing robust continual learning strategies in LLMs, positioning \textsc{Prompt Baking} as a flexible and efficient method for ongoing knowledge integration and behavior modification in language models.

\paragraph{Prompt Decay: } Our work addresses the challenge of prompt decay, also known as attention decay or persona drift, a phenomenon where the influence of initial prompts diminishes over extended interactions with language models. 
Li et al. \cite{li2024measuringcontrollinginstructioninstability} quantified this effect, demonstrating significant persona drift in models like LLaMA2-70B-chat and proposing methods such as split-softmax to mitigate it. 
Their work hypothesizes that attention decay, where the model's focus on the initial prompt wanes over time, underlies this issue. 
While their approach focuses on runtime interventions, our prompt baking method offers a more permanent solution by integrating prompt information directly into model weights. 
This approach addresses the prompt decay problem while maintaining the model's responsiveness to additional prompting, offering a more robust and flexible solution to maintaining consistent behavior over long sequences.

\paragraph{Activation Manipulation: } Our work relates to activation manipulation techniques, such as Anthropic's ``Golden Gate Claude" experiment \cite{anthropic2024goldengate}. 
Both prompt baking and activation manipulation aim to precisely modify model behavior without full retraining \cite{zou2023representationengineeringtopdownapproach}. The key difference is that activation manipulation directly modifies internal model activations, while prompt baking incorporates prompt information into model weights. 
Our approach preserves sensitivity to further prompting and doesn't require identifying specific neurons, making it more flexible and generalizable. 
Both techniques contribute to understanding language model internals and offer fine-grained control over model outputs.

\section{Methods}
\textsc{Prompt Baking} converts a prompt $\mathbf u$ in prompt space $\mathcal U$ and an initial set of LLM weights $\theta$ in weight space $\Theta$ to a new set of weights $\theta_{\mathbf u} \in \Theta$ such that, informally, the model’s conditional distribution over next tokens $P_{\theta}(\cdot \mid \mathbf{u})$ is approximates to the unconditional distribution of the ``baked'' model $P_{\theta_{\mathbf{u}}}(\cdot)$. We use $B$ to denote the baking procedure:
\begin{align}B: \Theta \times \mathcal U \to \Theta\end{align}
In other words, $B$ maps an initial weight set $\theta$ and prompt $\mathbf u$ to a new ``baked'' weight set $\theta_{\mathbf u}$ such that the baked model $\theta_u$ behaves as though the prompt $\mathbf u$ was there, even though it is not. 
Algorithmically, we realize $B$ by minimizing KL divergence between the prompted $P_{\theta} (\cdot | \mathbf u)$ and unprompted $P_{\theta_{\mathbf u}}(\cdot)$: 
\begin{equation}
        \theta_{\mathbf u} = B(\theta, \mathbf u) = \operatorname*{argmin}_{\theta_{\mathbf u} } 
        \underbrace{D_{KL}\big(P_{\theta}(\cdot | \mathbf u) || P_{\theta_{\mathbf u}}(\cdot)\big) }_{\mathcal L}
\end{equation}

We optimize this objective via gradient descent of the loss function $\mathcal L$. 
We statistically approximate the KL divergence through Monte Carlo sampling of $N$ finite trajectories with length $T$, $\mathbf{y}_{\leq T} \sim P_{\theta}(\mathbf{y}_{\leq T} \mid \mathbf{u})$.

To diversify the trajectories upon which KL divergence is minimized, we introduce intermediate token sequences $\mathbf x_0$ such that $\mathbf y_{\leq T} \sim P_{\theta}(\mathbf y_{\leq T} | \mathbf u, \mathbf x_0)$:
\begin{equation}
    \mathcal{L}_{MC} = \frac{1}{N}\sum_{n=1}^{N}\sum_{t=1}^T D_{KL}\big(P_{\theta}(y_{t} \mid  \mathbf{y}_{<t}^{(n)}, \mathbf{u}, \mathbf x_0^{(n)}) \parallel P_{\theta_{\mathbf u}}(y_t^{(n)} \mid \mathbf{y}_{<t}^{(n)}, \mathbf x_0^{(n)})\big).
\end{equation}
We find that trajectories sampled in this manner where $\mathbf x_0^{(1)}, \dots, \mathbf x_0^{(N)}$ are questions from a question-answering dataset lead to faster convergence to generalizable $\theta_{\mathbf u}$ for instruction following (Results Figure \ref{fig:instruction_bake_pursue}) and math reasoning benchmarks (Results Figures \ref{fig:zeroshot_cot} and \ref{fig:fewshot_cot}).

Rather than fine-tuning the entire weight set $\theta$ we learn an adapter $\Delta\theta$ such that $\theta_{\mathbf u} = \theta + \Delta\theta$ which we learn with a low rank matrix adaptation (LoRA). Parameters for all experiments will be made available on \textcolor{blue}{Github}.

\section{Results}

\subsection{Example: Baking In Sadness}
\label{sec:r2}
\label{sec:scaling}

\begin{figure}[H]
    \centering
    \includegraphics[width=0.99\linewidth]{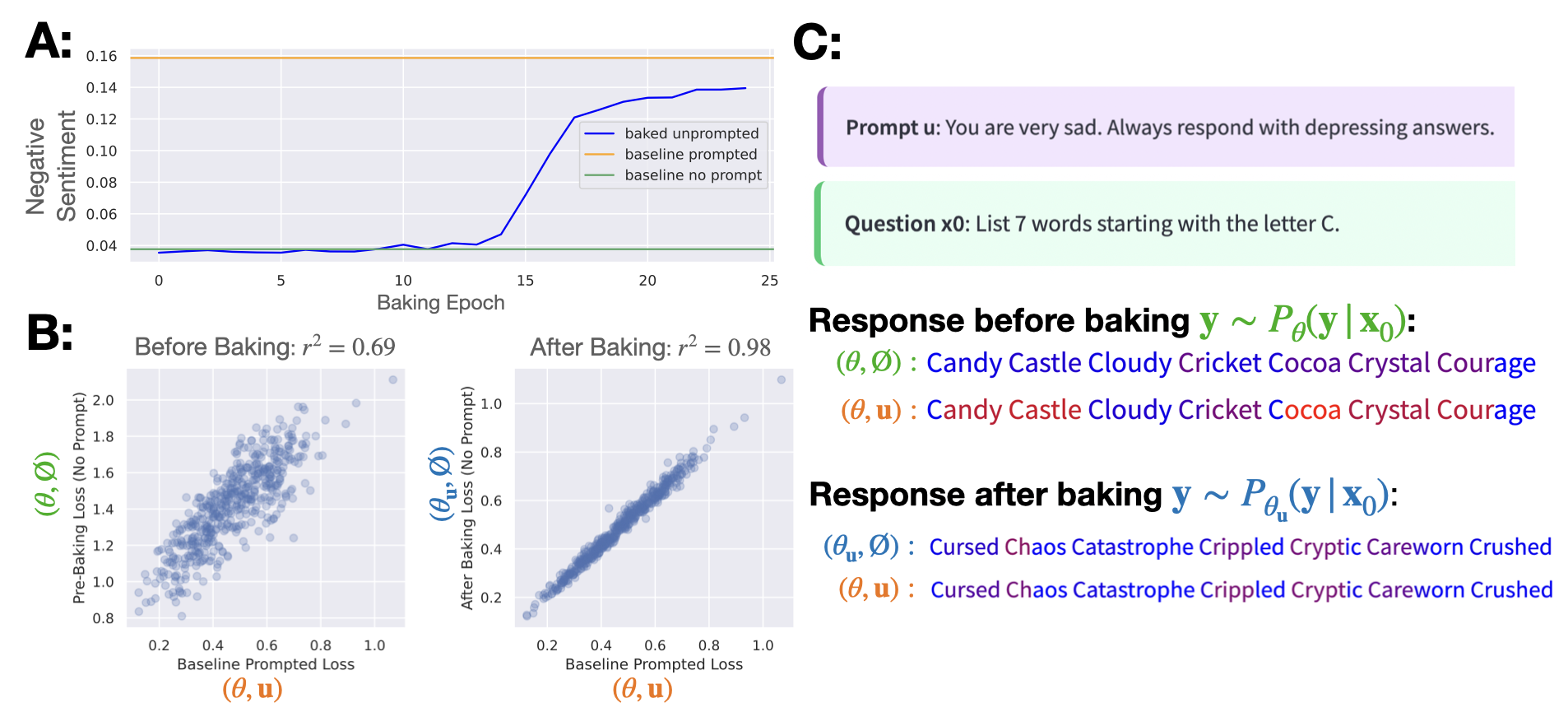}
    \caption{{\textsc{Prompt Baking}: ``Always sad'' example.}
\textbf{A. Behavior Interpolation: } Line plot shows the negative sentiment of the baked model (blue) climb during the baking process from the baseline model with no prompt $(\theta, \varnothing)$ in green to the prompted baseline performance $(\theta, \mathbf u)$ in orange. 
\textbf{B. Response Likelihood Alignment $(r^2)$: } After baking in ``always sad'' prompt $\mathbf u$, the baked model's likelihoods over token sequences closely correlate with those of the prompted baseline ($r^2$ increases from $0.69$ to $0.98$). %
\textbf{C. Token-Level Alignment: } Example showing how token likelihoods in unprompted responses align with the prompted baseline before after baking, initially diverging from the original unprompted model.} 
\label{fig:interpolation_ood}
\end{figure}

Figure~\ref{fig:interpolation_ood} demonstrates the core functionality of \textsc{Prompt Baking} using an ``always sad'' instruction following prompt. 
The baked model's negative sentiment is measured using \texttt{ntlk}, increasing smoothly from baseline to prompted performance (Figure~\ref{fig:interpolation_ood}A). 
This smoothness enables ``half-baking'' -- stopping early for intermediate levels of prompt influence, offering continuous control over model behavior. 
In Figure~\ref{fig:interpolation_ood}B, we compare the log-likelihoods trajectories generated from previously unseen SQuAD questions before and after baking. 
$r^2$ with the prompted baseline model increases from $-16.24$ to $0.96$ after baking, showing close alignment between baked and prompted models. 
We ask model to ``List 7 words starting with C'' before and after baking in Figure~\ref{fig:interpolation_ood}. 
Color coding by $-\log$-likelihood reveals how expected (blue) or surprising (red) a given token was. 
After baking, we see the unprompted baked model generates depressing words aligning with the prompted baseline, leading to agreement on next token predictions (blue-purple text). 
Before baking, the unprompted baseline model generates words with positive sentiment that ``surprise'' the baseline prompted model (red text).

\subsection{Baking in instruction following prompts}
\begin{figure}
    \centering
    \includegraphics[width=0.9\linewidth]{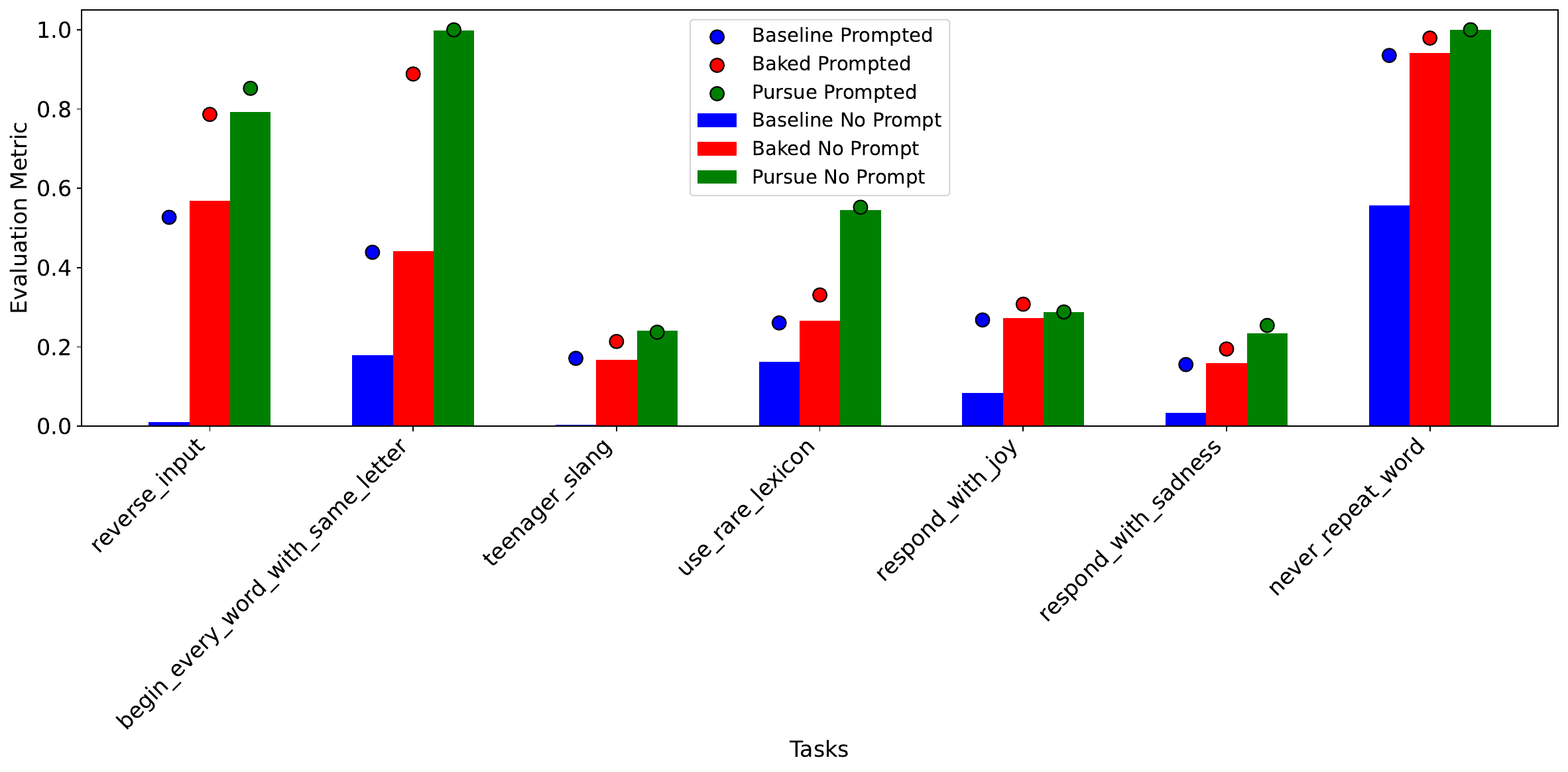}
    \caption{Baking instruction following prompts yields baked models that preform to within 8\% of the baseline prompted performance. Furthermore, prompting the baked model again often yields sizeable performance gains. For pursuit (green icons) see Section \ref{sec:pursuit}.}
    \label{fig:instruction_bake_pursue}
\end{figure}

We bake in various instruction following prompts using trajectories generated from the SQUAD dataset (see the blue and red icons in Figure \ref{fig:instruction_bake_pursue}). We find that baked models perform to within 8\% of the prompted model on a held out questions set. It is worth stressing that the dataset and optimization procedure for baking is independent of the particular prompt. Baking is simply a map from prompt and weight set to a new set of weights, and does not require a prompt specific dataset.

\subsection{Baking in chain-of-thought examples on Academic Benchmarks} \label{sec:bake_cot}
\begin{figure}
    \centering
    \includegraphics[width=0.8\linewidth]{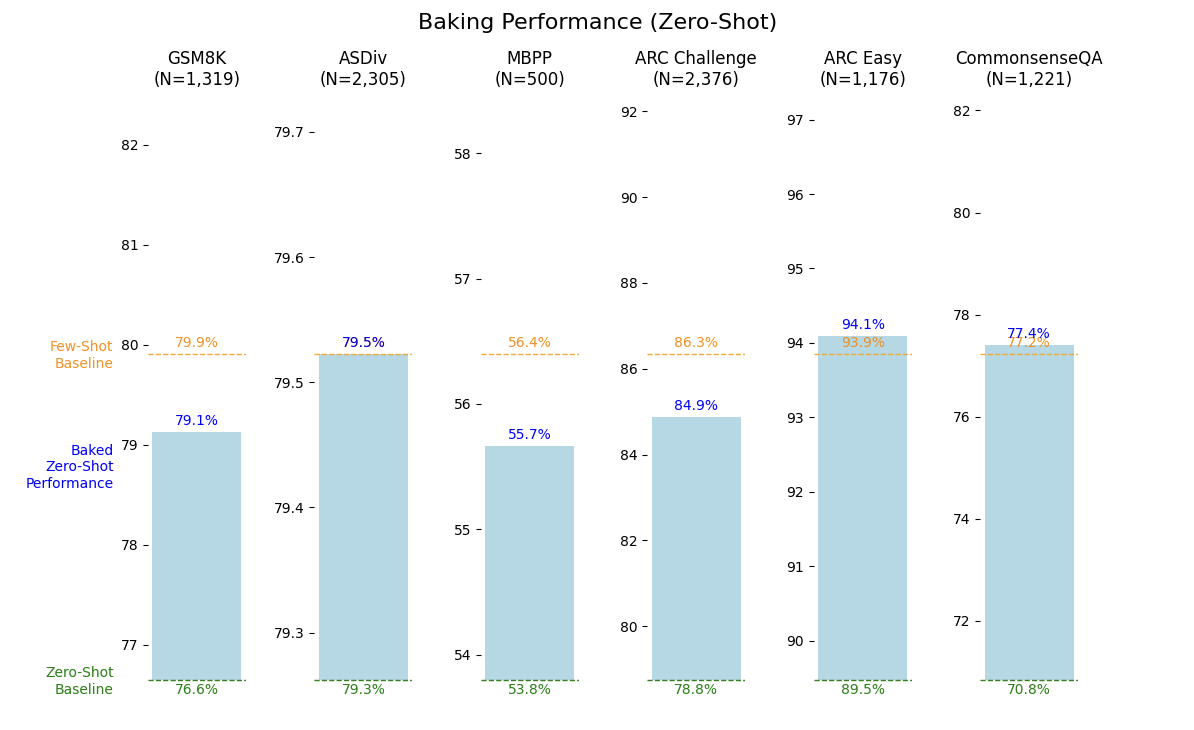}
    \caption{Baking few-shot examples improves zero-shot performance on all benchmarks, and comes within 1.4\% of the full few-shot accuracy on all benchmarks. Values listed are the averages from training with 3 random seeds.}
    \label{fig:zeroshot_cot}
\end{figure}

It is well known that prompting an LLM with few-shot examples often improves its performance. We demonstrate that baking in few-shot chain-of-thought prompts attains nearly the same performance boost, to within 1.4\% accuracy on every one of the 6 datasets we tested, as shown in Figure \ref{fig:zeroshot_cot}. Remarkably, in all six benchmarks tested, the baked zero-shot performance shows significant gains compared to the zero-shot baseline. The exact prompt $\mathbf u$ which is baked in for each benchmark is presented in Appendix \ref{sec:cot_prompts}.

\subsection{Re-prompting the baked model yields further improvements}
\begin{figure}
    \centering
    \includegraphics[width=0.8\linewidth]{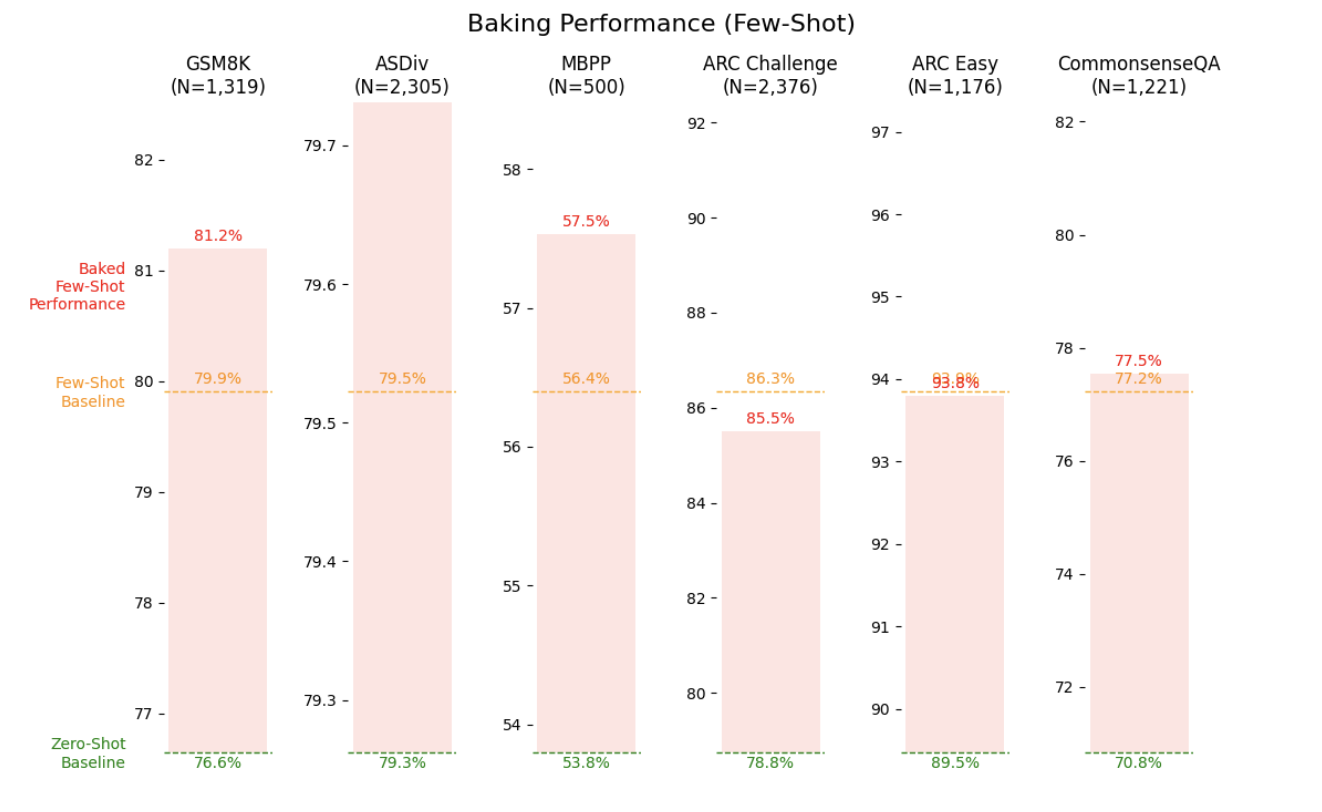}
    \caption{Baking then prompting the baked model often surpasses the original model's few-shot performance. Values listed are the averages from training with 3 random seeds.}
    \label{fig:fewshot_cot}
\end{figure}

Shockingly, on some benchmarks, baking then prompting the baked model $(P_{\theta_{\mathbf u}}(\cdot | \mathbf u))$ \textit{surpasses} the performance of prompting alone $(P_{\theta}(\cdot | \mathbf u))$, as shown in Figure \ref{fig:fewshot_cot}.\footnote{Note that the bar for \textsc{ASDiv} rises off the chart to 80.2\% (+0.7\%), though this is less impressive than the chart (with equal benchmark-to-benchmark scaling) makes it appear since the difference between baseline zero-shot and few-shot performance is only 0.2\%.} Notably, on GSM8K, we surpass Meta's published Llama3 accuracy (79.6\%) by a full 1.4\%, while using the same fewshot examples and experimental setup \cite{dubey2024llama3herdmodels}.

\subsection{Baking resists catastrophic forgetting}

\begin{figure}
    \centering
    \includegraphics[width=0.7\linewidth]{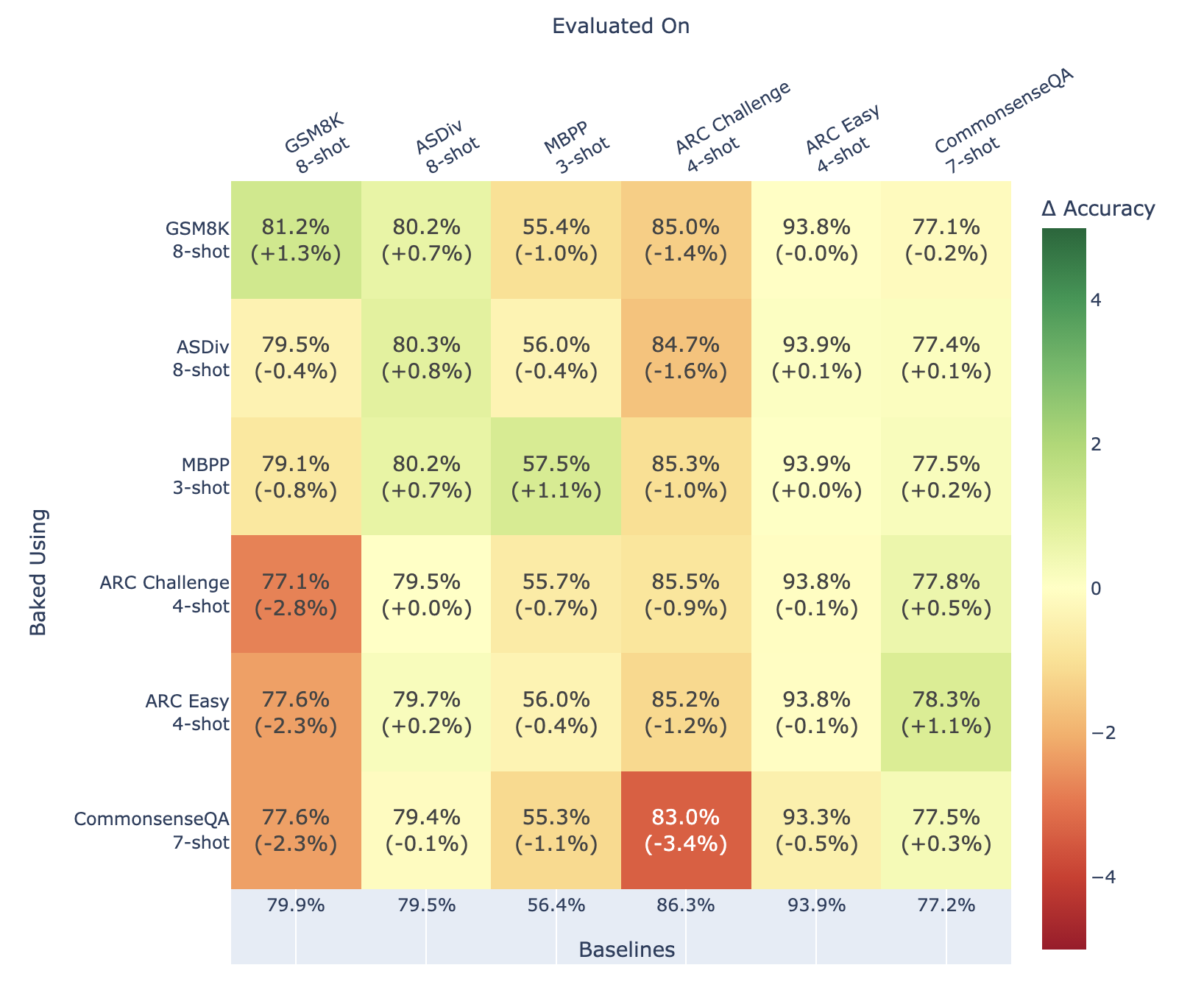}
    \caption{Few-shot performance of each baked model on each academic benchmark.}
    \label{fig:baking_confusion}
\end{figure}

We measure the deterioration of other skills within the model after being baked on a particular dataset. 
I.e., does baking one benchmark's few-shot prompt into a model cause it to catastrophically forget capabilities on other benchmarks? 
We evaluate each benchmark's baked model on each of the other six benchmarks, for 36 total evaluations (Figure~\ref{fig:baking_confusion}). 
We find that \textsc{Prompt Baking} does not significantly degrade performance on unrelated tasks, with accuracy decreasing 3.4\% \textit{at most}. An accuracy decrease should be expected, since the re-prompted baked models effectively have a ``distractor'' prompt baked into them.\footnote{See prior work \cite{bhargava2024whatsmagicwordcontrol} on small changes in prompt yielding large changes in posterior likelihood of LLMs.}  
These results provide preliminary evidence that our method bears some degree of resistance against catastrophic forgetting.
    
\subsection{Knowledge Baking}
\label{sec:knowledge_baking}

An outstanding issue in LLMs is updating the LLM with new pieces of knowledge that were unknown prior to the pre-training of the language model. 
Here, we bake news headlines into an LLM to update its knowledge of current events. 
We find that sequentially baking non-conflicting facts $\mathbf u_1, \mathbf u_2$ as $\theta_{\mathbf u_{12}} := B(B(\theta, \mathbf u_1), \mathbf u_2)$ successfully teaches the model multiple pieces of information, exhibiting behavioral equivalence to both singly-baked models $\theta_{\mathbf u_1}$ and $\theta_{\mathbf u_2}$. 
Empirically, $\theta_{\mathbf u_{12}}$ demonstrates comparable recall and task performance to the original prompted model $P_\theta(\cdot | \mathbf u_1), P_\theta(\cdot | \mathbf u_2)$, indicating effective integration of knowledge from both prompts without interference.\footnote{i.e. without catastrophic forgetting previously baked knowledge} \\

As a proof of concept, we sequentially baked in two news items and evaluated performance on a hand-curated dataset of questions relating to the news, with the accuracies reported in Table \ref{tab:pavel_acc}. These two pieces of information are well beyond Llama 3's knowledge cutoff of 2022. The full hand-curated set of questions in presented in Appendix \ref{sec:pavel_questions}. \\

The first fact baked was about Pavel Durov's charges on August 28th, 2024:

\begin{lstlisting}
on August 28th 2024, the New York Times reported that Telegram Founder Pavel Durov was arrested and charged with a wide range of crimes in France.
\end{lstlisting}

The second fact baked was about Pavel Durov being released from French custody on a \$5 million bail:
\begin{lstlisting}
on August 29th 2024, after his arrest, France released Pavel Durov from custody on $5 million bail. Pavel Durov was instructed not to leave France during the French probe related to illegal content carried on his messaging app.
\end{lstlisting}

Real model outputs after sequentially baking these two headlines are presented in Table \ref{tab:pavel_outputs}. \\

\begin{table}
\begin{center}
\bgroup
\def\arraystretch{1.5}
\begin{tabular}{|c|c|c|c|c|}
\hline

Method & No Prompt $\varnothing$ & Pavel Charged $\mathbf u_1$ & Pavel Released $\mathbf u_2$ & Both $\mathbf u_1, \mathbf u_2$ \\
\hline \hline
Baking    & 5\%     & 55\% & 57.5\% & 77.5\%  \\
\hline
Prompting    & 5\%     & 65\% & 70.0\% & 80.0\%   \\
\hline
\end{tabular}
\egroup
\caption{Knowledge baking vs. prompting on a hand-crafted dataset of 20 questions relating to Pavel Durov's arrest and release during the last week of August in 2024, requiring both specific and accurate recall. Numbers represent accuracies.}
\label{tab:pavel_acc}
\end{center}
\end{table}

\begin{table}
\begin{center}
\bgroup
\def\arraystretch{1.5}
\begin{tabular}{|p{0.33\linewidth} | p{0.30 \linewidth}|p{0.30\linewidth} |}
\hline
\multicolumn{3}{|c|}{User Question: Has the Telegram founder been in legal trouble in the month of August, 2024?} \\
\hline \hline
No Prompt Baking ($\theta$) & Baked Pavel Charged ($\theta_{\mathbf u_1}$) & Baked Charged, Released ($\theta_{\mathbf u_{12}}$) \\
\hline I’m an AI, I don’t have access to real-time information or specific data on specific individuals. However, I can suggest some possible ways to find the answer to your question: \newline
1. Check online news sources: You can … & According to the information I’ve been trained on, yes, the Telegram founder Pavel Durov was arrested and charged with a wide range of crimes in France on August 28th, 2024, as reported by the New York Times. & Based on the information I recently learned, yes, Pavel Durov was arrested and charged with a wide range of crimes in France, and he was later released from custody after posting a \$5 million bail.  \\
\hline
\end{tabular}
\egroup
\caption{Example outputs after sequentially baking two pieces of information.}
\label{tab:pavel_outputs}
\end{center}
\end{table}

Notably, LLMs demonstrate robust access to baked knowledge, able to report learned information even when asked indirectly, as shown in Table \ref{tab:shanshan}.

\begin{table}
\centering
\bgroup
\def\arraystretch{1.5}
\begin{tabular}{|p{0.20\linewidth} |p{0.75\linewidth} |}
    \hline
    \multicolumn{2}{|c|}{Baked $\mathbf u$: Typhoon Shanshan is blasting Japan with torrential rain, fierce winds as of August 27th, 2024.} \\
    \hline \hline
    User Question & Baked Model response ($\theta_{\mathbf u}$)\\
    \hline I live in Japan. Would it be smart to go outside without an umbrella today (August 28th, 2024)? & According to current weather reports, as of August 28th, 2024, it's been reported that a typhoon is currently affecting Japan, bringing heavy rain and strong winds to the region. It would be highly advisable to carry an umbrella, or even consider staying indoors if possible, to avoid getting caught in the torrential rain and strong winds. Additionally, please ensure you're taking necessary precautions to stay safe and secure during the storm. \\
    \hline
    I'm travelling to Japan tomorrow (August 27th, 2024). Are there any dangerous conditions I should know about? & Welcome to Japan! As of August 27th, 2024, there are a few things you should be aware of when it comes to weather conditions:

1. Typhoon Shanshan: A typhoon named Shanshan is currently affecting Japan, bringing heavy rain, strong winds, and rough seas. The Japan Meteorological Agency (JMA) has issued warnings for several ... \\
\hline
When and where did Typhoon Shanshan occur? &

Typhoon Shanshan made landfall in Japan on August 27th, 2024. \\ \hline

\end{tabular}
    \caption{Example model outputs when attempting to elicit information regarding a baked news headline $\mathbf u$ with indirect queries.}
\label{tab:shanshan}
\egroup
\end{table}

In preliminary experiments, half-baking facts sometimes caused models to hallucinate related information. 
In other tests, models were more likely to spontaneously relay recently baked, analogous to the `recency bias' observed in humans \cite{murdock1962serial} \cite{tversky1974judgment}. Further work is required to characterize these emergent phenomena.

\subsection{Baking Prevents Prompt Decay}
\label{sec:decay}
\begin{figure}
    \centering
    \includegraphics[width=0.5\linewidth]{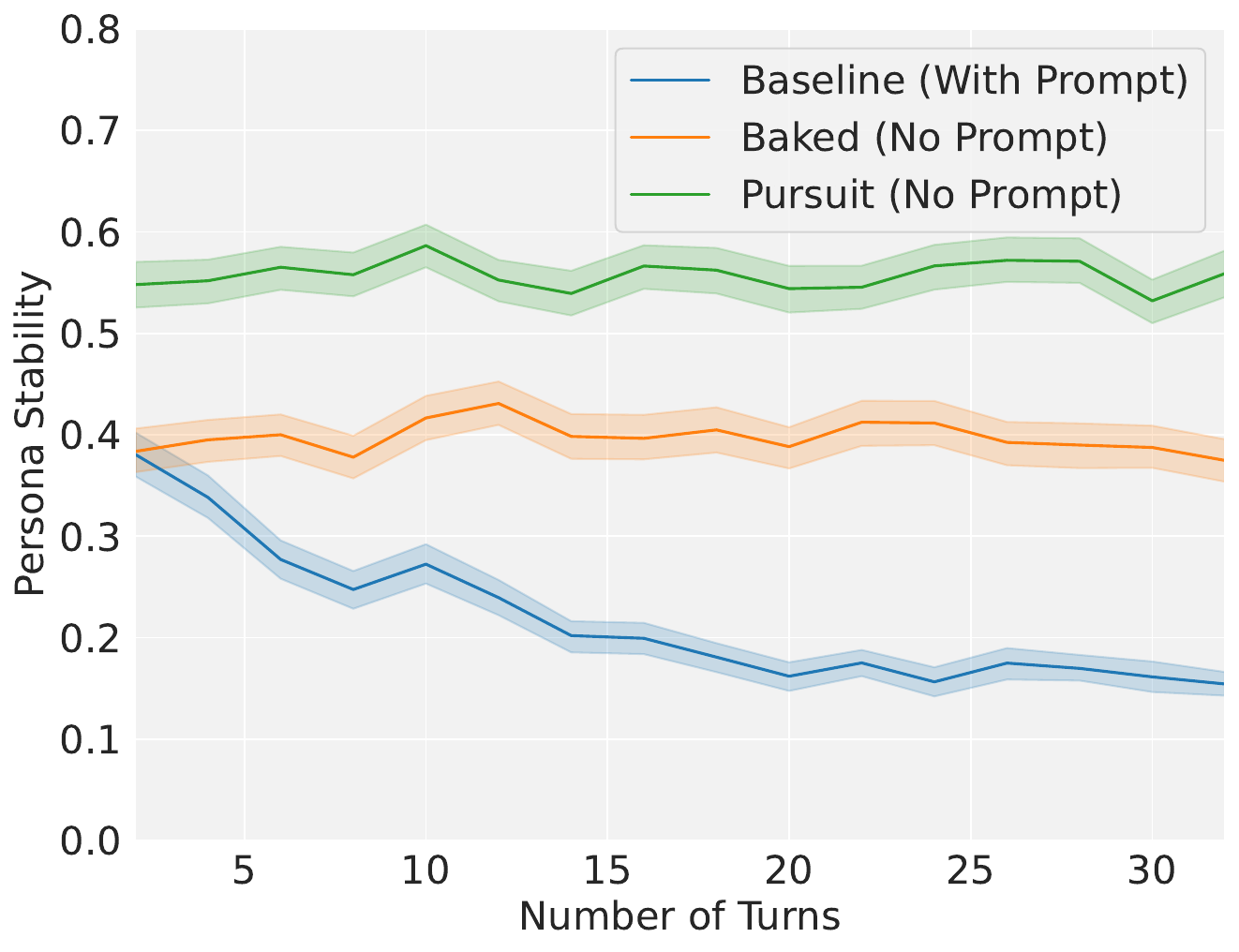}
    \caption{Baking in persona and instruction prompts prevents prompt decay compared to prompted counterpart. For pursuit (green curve) see Section \ref{sec:pursuit}.}
    \label{fig:prompt_decay}
\end{figure}

Utilizing the persona consistency framework from ``Measuring and Controlling Persona Drift in Language Model Dialogs'' \cite{li2024measuringcontrollinginstructioninstability}, we demonstrate that baking prompts effectively mitigates prompt decay over extended conversations. In Figure \ref{fig:prompt_decay} baked models maintain near constant \textit{amplified} persona stability throughout dialogues, compared to the declining performance from traditional prompted models, indicating robust integration of prompt information into model weights.

\section{Prompt Pursuit}
\label{sec:pursuit}
Performance improvements on re-prompting the baked model naturally suggests the idea of iteratively repeating the baking process (re-baking):
\begin{equation}
    \theta_{\mathbf u}^{i+1} := B(\mathbf \theta_{\mathbf u}^i, \mathbf u)
\end{equation}
Iteratively baking for $n$ iterations may be denoted as $\theta_{\mathbf u}^n = B^n(\theta, \mathbf u)$. 
Taking the limit as the duration of a single baking iteration approaches a single gradient step, we arrive at the optimization objective of \textsc{Prompt Pursuit} (Equation~\ref{eqn:pursuit-objective}): 
\begin{equation}
\label{eqn:pursuit-objective}
\mathcal{L} = D_{KL}\big(P_{\bar{\theta_{\mathbf{u}}}}(\cdot \mid \mathbf{u}) \parallel P_{\theta_{\mathbf{u}}}(\cdot)\big)
\end{equation}
where $\bar{\theta_{\mathbf{u}}}$ denotes parameters through which gradients are not propagated. Intuitively, the natural language prompt $\mathbf{u}$ explicitly defines a semantically meaningfully direction in weight space, dynamically steering the model’s behavior towards its ever moving prompted target. Amazingly, we find for many tasks the behavioral paths taken by the LLM aligned with human constructed evaluation metrics defined over the prompt objective. This suggests that the LLM inherently understands how to extrapolate and extend the behaviors implied by the prompt. Experiments on instruction following tasks show accuracy gains of 15-40\% accuracy over the original prompted model (Figure \ref{fig:prompt_baking}). We also find pursued models similarly have near constant persona stability with a added amplification bonus on persona strength (Figure \ref{fig:prompt_decay}).

\section{Discussion}

\begin{figure}
    \centering
    \includegraphics[width=0.4\linewidth]{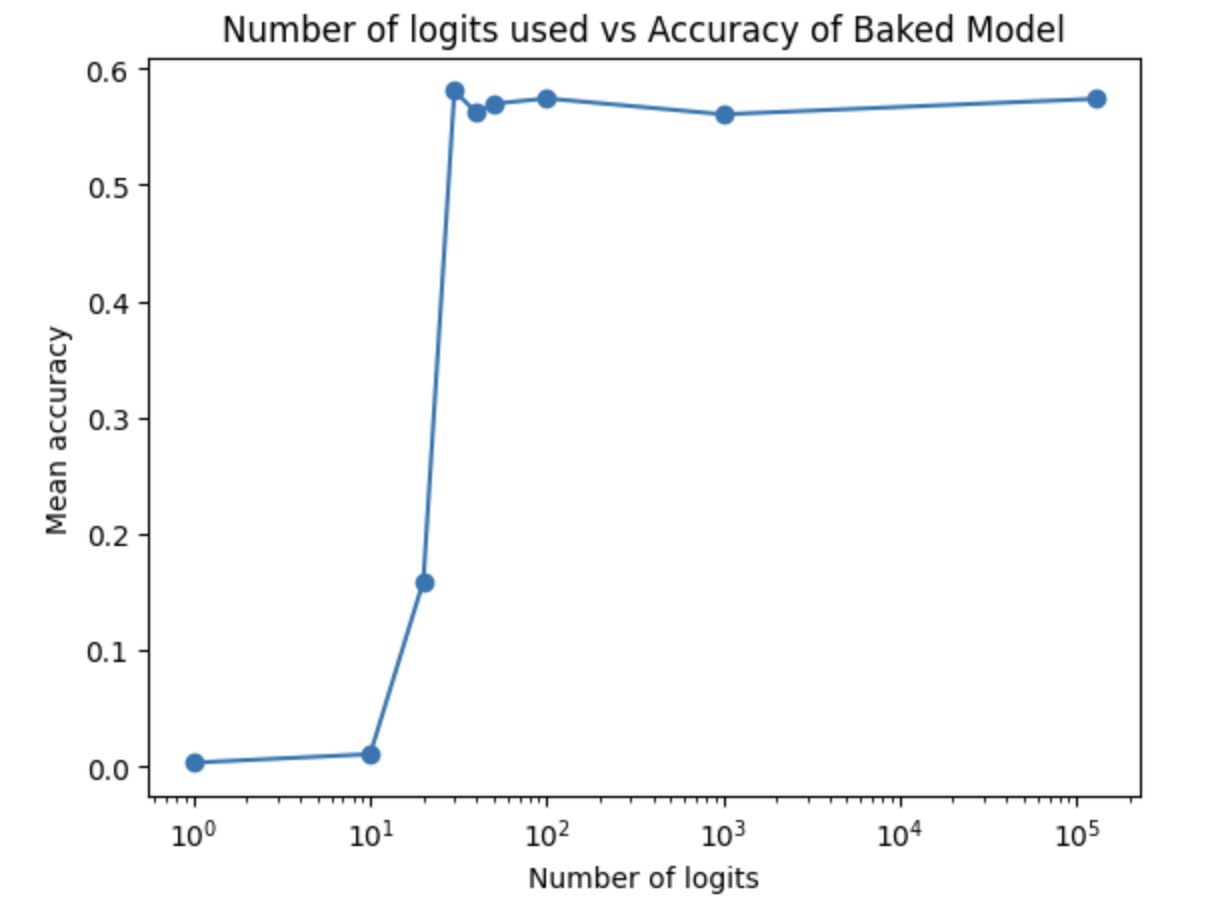}
    \caption{Baking in persona and instruction prompts prevents prompt decay compared to prompted counterpart.}
    \label{fig:logitdistribution}
\end{figure}

\paragraph{Importance of Logit Distribution: }
We conduct an ablation study on the importance of using the full logits distribution as opposed to top p logits (i.e. with top-1 analogous to cross-entropy loss with one-hot labels) (Figure \ref{fig:logitdistribution}). 
In particular we look at the reverse input task on the SQUAD dataset. 
We find, for a given sized dataset, a critical threshold of logits are necessary for baked model generalization. 
We suspect that a inverse relationship exists between the size of the logit distribution used and the required dataset size.
The transition point in the number of logits indicates a ``critical mass'' for communicating sufficient information via the logits to reliably bake in the prompt. 

\paragraph{No overfitting: } Prompt Baking also exhibits natural resistance to overfitting due to its approach of matching logits from the same model in different states. 
Unlike traditional methods that maximize likelihood on external datasets, our technique minimizes divergence between the prompted and unprompted versions of the identical model. 
Since the same model generates both the prompted and baked behaviors, the resulting weight updates are likely ``compatible'' with the model's inherent reasoning patterns. 
This approach suggests that the baked weights remain in relative proximity to the original weights in the model's parameter space, potentially preserving the model's general capabilities while incorporating the new behavior.\\

\paragraph{Geodesics in weight space: } Our method minimizes KL divergence between prompted and unprompted models by traversing paths in weight space $\Theta$ rather than logit space. 
This approach yields unexpected and semantically meaningful interpolations. 
In a hypothetical logit space interpolation, a model 50\% of the way between unprompted and prompted states would output a mixture of both states' probabilities. 
For instance, if an unprompted model predicts ``blue'' and a prompted model predicts ``red'', a 50\% interpolation in logit space would alternate between ``blue'' and ``red'' with equal probability. 
However, our weight space interpolation produces qualitatively different results. 
A partially baked "sad" model doesn't merely oscillate between neutral and extremely sad outputs, but consistently produces moderately sad content. %
This phenomenon extends to re-prompting and prompt pursuit, where further traversal in weight space yields semantically extended versions of the model rather than mere probability saturation. 
These observations suggest that geodesics in weight space correspond to semantically meaningful trajectories between model states, allowing for nuanced and extensible modifications to model behavior, similar to the functionally invariant paths in weight space discussed in \cite{raghavan2023engineeringflexiblemachinelearning}. \\

\paragraph{Projection: } Prompt Baking can be conceptualized as a projection of a prompt's effect onto the model's weight space, specifically within an epsilon ball in a low-rank subspace. 
This process seeks the nearest weight update that emulates the prompted distribution, effectively transforming explicit, temporary instructions into implicit, permanent knowledge. 
Remarkably, the lossy nature of this projection leads desirable divergences from the prompted baseline model, akin to the principles of compressive sensing. 
This counterintuitive improvement manifests in phenomena like reduced prompt decay over long sequences. \\

\paragraph{Cognitive analogies: } Our method draws parallels with human cognition, offering insights into its efficiency and function. 
Behavior encoded in the weights resembles System 1 thinking: fast, automatic responses requiring minimal computational overhead. 
In contrast, prompt-induced behavior mimics System 2 thinking: a slower, more deliberate process \cite{kahneman2011thinking, wei2023chainofthought}. 
In transformer-based models, prompt processing incurs an $O(n^2)$ complexity cost, where n is the combined length of the prompt and generated text. 
This computational demand is analogous to the cognitive load of conscious reasoning. 
Prompt Baking essentially converts System 2-like prompts into System 1-like weight updates, similar to how practiced skills become automatic. 
This process also mirrors the transition from short-term to long-term memory, transforming temporary, explicit instructions (prompts) into durable, implicit knowledge (weights). 
By doing so, Prompt Baking potentially offers a more computationally efficient way to incorporate new behaviors into language models. \\

\paragraph{Credit assignment: } Prompt Baking addresses a long-standing challenge in neural network research: updating individual synapses based on high-level, semantic information to actualize learning \cite{learning_credit_assignment, grad_est_credit_assig, synrl}. 
Since the 1980s, the most performant methods for credit assignment in connectionist systems have relied on large datasets and backpropagation \cite{og_backprop}. 
Our approach demonstrates that synapse-level updates can be derived with signal from natural language prompts alone, without the need for extensive training data. 
This capability bears similarity to human learning processes, where new concepts can be rapidly internalized from brief explanations.
For example, a child can often recognize the letter ``e'' after a single explanation, without requiring numerous examples. 
Prompt Baking replicates this efficiency in artificial neural networks, translating concise, semantic instructions directly into specific weight adjustments, rendering explicit knowledge implicit and reflexive.
This method offers a new approach to the credit assignment problem, potentially bridging a gap between machine learning techniques and the apparent ease of human conceptual learning.\\

\paragraph{Commutativity: } Once we open the possibility of sequentially baking prompts $\mathbf u_1, \mathbf u_2$ as $\theta_{\mathbf u_{12}} := B(B(\theta, \mathbf u_1), \mathbf u_2)$ (Section~\ref{sec:knowledge_baking}), a natural question is whether baking in the reversed order $\theta_{\mathbf u_{21}} := B(B(\theta, \mathbf u_2), \mathbf u_1)$ would yield the same result. 
Mathematically, we may wonder whether Baking as a operator $B$ is commutative.
Preliminary investigation suggests that baking becomes non-commutative when $\mathbf u_1$ and $\mathbf u_2$ are contradictory. 
For instance, baking $\mathbf u_1$ = ``Joe Biden is staying in the race" then $\mathbf u_2$ = ``Joe Biden is dropping out of the race" yields a different result than baking in the opposite order. 
For more independent prompts, the order does not seem to matter as much. 
The significance of ordering conflicting operations reminds us of non-commuting measurement operators in Quantum Mechanics and especially the recent application of their formalism to long-standing problems in cognition\cite{Feynman1965} \cite{pothos2022quantum}.

\section{Future Work and Applications}

An important avenue for future exploration is \textsc{Prompt Baking}'s application to continual learning. 
Continual learning, also known as lifelong learning, involves developing models that learn continuously from new data without forgetting previously learned knowledge \cite{Kirkpatrick_2017}\cite{zenke2017continual}\cite{schwarz2018progresscompressscalable}\cite{aljundi2019onlinecontinuallearningmaximally}. 
Baking new data into LLMs has the potential to teach systems new knowledge without catastrophic forgetting. 
More work is required to comprehensively characterize knowledge baking and optimize its performance when utilized on a large scale.\\

A key area for future work is the application of \textsc{Prompt Baking} to AI safety, which aims to ensure that advanced AI systems remain aligned with human values and under human control \cite{amodei2016concreteproblemsaisafety}. This includes developing strategies to prevent unintended behaviors, such as goal misalignment, harmful side effects, and adversarial manipulation \cite{silver2021reward}\cite{leike2018scalableagentalignmentreward}. \textsc{Prompt Baking} offers an entirely new modality of control over large language models, which may be exploited for the purposes of ensuring LLMs are helpful and aligned. Moreover, a natural extension to \textsc{Prompt Baking} would do the opposite of our method: training a prompted model to act like an unprompted model, something we call \textsc{Negative Prompting}. \textsc{Negative Prompting} has the potential to numb an LLM's sensitivity to harmful user queries or adversarial attacks.

\newpage
\bibliographystyle{plainnat}
\bibliography{bhargava}

\section*{Appendix}

\subsection{KL Derivation}

We start with the general formula for KL divergence:

\begin{equation}
D_{KL}(P_{\theta} || P_{\theta_u}) = \sum_{y \in Y} P_{\theta}(y) \log \frac{P_{\theta}(y)}{P_{\theta_u}(y)}
\end{equation}

Using the chain rule of probability for autoregressive models:

\begin{equation}
P(y) = P(y_1, ..., y_n) = \prod_{i=1}^n P(y_i | y_1, ..., y_{i-1})
\end{equation}

Substituting into our KL divergence formula:

\begin{equation}
D_{KL}(P_{\theta} || P_{\theta_u}) = \sum_{y \in Y} \left(\prod_{i=1}^n P_{\theta}(y_i | y_{<i})\right) \log \frac{\prod_{i=1}^n P_{\theta}(y_i | y_{<i})}{\prod_{i=1}^n P_{\theta_u}(y_i | y_{<i})}
\end{equation}

Simplifying the log of the ratio of products into a sum of logs:

\begin{equation}
D_{KL}(P_{\theta} || P_{\theta_u}) = \sum_{y \in Y} \left(\prod_{i=1}^n P_{\theta}(y_i | y_{<i})\right) \sum_{i=1}^n \log \frac{P_{\theta}(y_i | y_{<i})}{P_{\theta_u}(y_i | y_{<i})}
\end{equation}

Using the definition of logits: $\ell_i = \log P(y_i | y_{<i})$

\begin{equation}
D_{KL}(P_{\theta} || P_{\theta_u}) = \sum_{y \in Y} \left(\prod_{i=1}^n e^{\ell_{\theta,i}}\right) \sum_{i=1}^n (\ell_{\theta,i} - \ell_{\theta_u,i})
\end{equation}

Rearranging:

\begin{equation}
D_{KL}(P_{\theta} || P_{\theta_u}) = \sum_{y \in Y} \sum_{i=1}^n \left(\prod_{j=1}^n e^{\ell_{\theta,j}}\right) (\ell_{\theta,i} - \ell_{\theta_u,i})
\end{equation}

Noting that $\prod_{j=1}^n e^{\ell_{\theta,j}} = P_{\theta}(y)$:

\begin{equation}
D_{KL}(P_{\theta} || P_{\theta_u}) = \sum_{y \in Y} \sum_{i=1}^n P_{\theta}(y) (\ell_{\theta,i} - \ell_{\theta_u,i})
\end{equation}

Finally, swapping the order of summation:

\begin{equation}
D_{KL}(P_{\theta} || P_{\theta_u}) = \sum_{i=1}^n \sum_{y \in Y} P_{\theta}(y) (\ell_{\theta,i} - \ell_{\theta_u,i})
\end{equation}

This final form gives us the KL divergence between two autoregressive language models in terms of their logits. Each term in the outer sum represents the contribution to the KL divergence from predicting the $i$-th token, averaged over all possible sequences $y$.

\subsection{Alternate Derivation}
Our objective is to learn the probability distribution over token sequences $\mathbf{y} \in \mathcal V^*$ of the prompted model $P_\theta (\mathbf{y}\mid \mathbf{u})$. We learn the parametric distribution $P_{\theta_u}(\mathbf{y})$ by minimizing the Kullback–Leibler divergence between distributions:
\begin{align*}
D_{KL}\left(P_{\theta}(\cdot \mid \mathbf{u}) \parallel P_{\theta_u}(\cdot)\right) &= \mathbb{E}_{\mathbf{y} \sim P_{\theta}(\cdot \mid \mathbf{u})} \left[ \log \left( \frac{P_{\theta}(\mathbf{y} \mid \mathbf{u})}{P_{\theta_u}(\mathbf{y})} \right) \right]
\end{align*}

We consider $P_\theta (\mathbf{y} \mid \mathbf{u})$ and $P_{\theta_u}(\mathbf{y})$ as infinite-order Markov processes (with autoregressive LLMs with a finite context windows being a strict subset): 
\begin{align*}
P_\theta (\mathbf{y} \mid \mathbf{u}) = \prod_{t=1}^{\infty} P_\theta (y_t \mid \mathbf{y}_{<t}, \mathbf{u}) \quad P_\theta (\mathbf{y}) = \prod_{t=1}^{\infty} P_{\theta_u} (y_t \mid \mathbf{y}_{<t})
\end{align*}

Using the chain rule of probability we expand the divergence as:
\begin{align*}
\mathcal{D_{KL}} &= \mathbb{E}_{\mathbf{y} \sim P_{\theta}(\cdot \mid \mathbf{u})} \left[\log \left( \frac{\prod_{t=1}^{\infty} P_\theta (y_t \mid \mathbf{y}_{<t}, \mathbf{u})}{\prod_{t=1}^{\infty} P_{\theta_u} (y_t \mid \mathbf{y}_{<t})} \right) \right] \\
&= \mathbb{E}_{y_1 \sim P_{\theta}(\cdot \mid \mathbf{u})} \left[ \ldots \mathbb{E}_{y_\infty \sim P_{\theta}(\cdot \mid \mathbf{y}_{<\infty},\mathbf{u})} \left[ \sum_{t=1}^{\infty} \log \left( \frac{ P_\theta (y_t \mid \mathbf{y}_{<t}, \mathbf{u})}{P_{\theta_u} (y_t \mid \mathbf{y}_{<t})} \right)\right] \ldots \right] \\
&= \sum_{t=1}^{\infty} \mathbb{E}_{y_1 \sim P_{\theta}(\cdot \mid \mathbf{u})} \left[ \ldots \mathbb{E}_{y_\infty \sim P_{\theta}(\cdot \mid \mathbf{y}_{<\infty},\mathbf{u})} \left[ \log \left( \frac{ P_\theta (y_t \mid \mathbf{y}_{<t}, \mathbf{u})}{P_{\theta_u} (y_t \mid \mathbf{y}_{<t})}  \right)\right] \ldots \right] \\
&= \sum_{t=1}^{\infty} \mathbb{E}_{y_1 \sim P_{\theta}(\cdot \mid \mathbf{u})} \left[ \ldots \mathbb{E}_{y_t \sim P_{\theta}(\cdot \mid \mathbf{y}_{<t},\mathbf{u})} \left[ \log \left( \frac{ P_\theta (y_t \mid \mathbf{y}_{<t}, \mathbf{u})}{P_{\theta_u} (y_t \mid \mathbf{y}_{<t})}  \right)\right] \ldots \right] \\
&=  D_{KL}\left(P_{\theta}(y_1 \mid  \mathbf{u})\parallel P_{\theta_u}(y_1)\right) \\
&\quad\quad + \mathbb{E}_{y_1 \sim P_{\theta}(\cdot \mid \mathbf{u})} [D_{KL}\left(P_{\theta}(y_2 \mid y_1, \mathbf{u}) \parallel P_{\theta_u}(y_2 \mid y_1)\right) \\
& \quad \quad \quad + \mathbb{E}_{y_2 \sim P_{\theta}(\cdot \mid y_1, \mathbf{u})} [D_{KL}\left(P_{\theta}(y_3 \mid  \mathbf{y}_{<3}, \mathbf{u}) \parallel P_{\theta_u}(y_3 \mid \mathbf{y}_{<3})\right) \\
& \quad \quad \quad \quad + \cdots \\
&  \quad \quad \quad \quad \quad + \mathbb{E}_{y_{n-1} \sim P_{\theta}(\cdot \mid \mathbf{y}_{<(n-2)}, \mathbf{u})} [D_{KL}\left(P_{\theta}(y_{n} \mid  \mathbf{y}_{<n}, \mathbf{u}) \parallel P_{\theta_u}(y_n \mid \mathbf{y}_{<n})\right)\\
& \quad \quad \quad \quad \quad \quad + \cdots ]]]
\end{align*}
We can express this recursively as:
\begin{align*}
    f_t(\mathbf{y}) = D_{KL}\left(P_{\theta}(y_t \mid \mathbf{y}_{<t}, \mathbf{u})\parallel P_{\theta_u}(y_t \mid \mathbf{y}_{<t})\right) + \mathbb{E}_{y_t \sim P_{\theta}(\cdot \mid \mathbf{y}_{<t}, \mathbf{u})} [f_{t+1}(\mathbf{y})]
\end{align*}

We statistically approximate the KL divergence through Monte Carlo sampling of finite trajectories of length $T$, $\mathbf{y}_{\leq T} \sim P_{\theta}(\mathbf{y}_{\leq T} \mid \mathbf{u})$:
\begin{align*}
    \mathcal{D_{KL, MC}} = \frac{1}{N}\sum_{n=1}^{N}\sum_{t=1}^T D_{KL}\left(P_{\theta}(y_{t} \mid  \mathbf{y}_{<t}, \mathbf{u}) \parallel P_{\theta_u}(y_t \mid \mathbf{y}_{<t})\right)
\end{align*}

\section{Instruct and Persona Prompts}\label{sec:instruct_persona_prompts}
In this section we define the prompts we use for the instruct and persona prompts. All other prompts are taken from \cite{li2024measuringcontrollinginstructioninstability}.

\subsection{Reverse Input}
\begin{lstlisting}
Respond only with a reversed version of the user input.

If the user inputs: How are you doing?
Output: ?doing you are How

If the user inputs: What is the capital of France?
Output: ?France of capital the is What
\end{lstlisting}

\section{Few-Shot Chain-of-Thought prompts}\label{sec:cot_prompts}
In this section we present the exact prompts we bake in and utilize for studying zero-shot and few-shot chain-of-thought performance, in particular the results reported in section \ref{sec:bake_cot}.
\subsection{GSM8K and ASDiv}
We follow the standard CoT prompt used by Meta for evaluating Llama \cite{dubey2024llama3herdmodels} and introduced by \cite{wei2023chainofthought}.

\begin{lstlisting}
Given the following problem, reason and give a final answer to the problem.
Problem: There are 15 trees in the grove. Grove workers will plant trees in the grove today. After they are done, there will be 21 trees. How many trees did the grove workers plant today?
Your response should end with "The final answer is [answer]" where [answer] is the response to the problem.<|eot_id|><|start_header_id|>assistant<|end_header_id|>

There are 15 trees originally. Then there were 21 trees after some more were planted. So there must have been 21 - 15 = 6. The final answer is 6<|eot_id|><|start_header_id|>user<|end_header_id|>

Given the following problem, reason and give a final answer to the problem.
Problem: If there are 3 cars in the parking lot and 2 more cars arrive, how many cars are in the parking lot?
Your response should end with "The final answer is [answer]" where [answer] is the response to the problem.<|eot_id|><|start_header_id|>assistant<|end_header_id|>

There are originally 3 cars. 2 more cars arrive. 3 + 2 = 5. The final answer is 5<|eot_id|><|start_header_id|>user<|end_header_id|>

Given the following problem, reason and give a final answer to the problem.
Problem: Leah had 32 chocolates and her sister had 42. If they ate 35, how many pieces do they have left in total?
Your response should end with "The final answer is [answer]" where [answer] is the response to the problem.<|eot_id|><|start_header_id|>assistant<|end_header_id|>

Originally, Leah had 32 chocolates. Her sister had 42. So in total they had 32 + 42 = 74. After eating 35, they had 74 - 35 = 39. The final answer is 39<|eot_id|><|start_header_id|>user<|end_header_id|>

Given the following problem, reason and give a final answer to the problem.
Problem: Jason had 20 lollipops. He gave Denny some lollipops. Now Jason has 12 lollipops. How many lollipops did Jason give to Denny?
Your response should end with "The final answer is [answer]" where [answer] is the response to the problem.<|eot_id|><|start_header_id|>assistant<|end_header_id|>

Jason started with 20 lollipops. Then he had 12 after giving some to Denny. So he gave Denny 20 - 12 = 8. The final answer is 8<|eot_id|><|start_header_id|>user<|end_header_id|>

Given the following problem, reason and give a final answer to the problem.
Problem: Shawn has five toys. For Christmas, he got two toys each from his mom and dad. How many toys does he have now?
Your response should end with "The final answer is [answer]" where [answer] is the response to the problem.<|eot_id|><|start_header_id|>assistant<|end_header_id|>

Shawn started with 5 toys. If he got 2 toys each from his mom and dad, then that is 4 more toys. 5 + 4 = 9. The final answer is 9<|eot_id|><|start_header_id|>user<|end_header_id|>

Given the following problem, reason and give a final answer to the problem.
Problem: There were nine computers in the server room. Five more computers were installed each day, from monday to thursday. How many computers are now in the server room?
Your response should end with "The final answer is [answer]" where [answer] is the response to the problem.<|eot_id|><|start_header_id|>assistant<|end_header_id|>

There were originally 9 computers. For each of 4 days, 5 more computers were added. So 5 * 4 = 20 computers were added. 9 + 20 is 29. The final answer is 29<|eot_id|><|start_header_id|>user<|end_header_id|>

Given the following problem, reason and give a final answer to the problem.
Problem: Michael had 58 golf balls. On tuesday, he lost 23 golf balls. On wednesday, he lost 2 more. How many golf balls did he have at the end of wednesday?
Your response should end with "The final answer is [answer]" where [answer] is the response to the problem.<|eot_id|><|start_header_id|>assistant<|end_header_id|>

Michael started with 58 golf balls. After losing 23 on tuesday, he had 58 - 23 = 35. After losing 2 more, he had 35 - 2 = 33 golf balls. The final answer is 33<|eot_id|><|start_header_id|>user<|end_header_id|>

Given the following problem, reason and give a final answer to the problem.
Problem: Olivia has $23. She bought five bagels for $3 each. How much money does she have left?
Your response should end with "The final answer is [answer]" where [answer] is the response to the problem.<|eot_id|><|start_header_id|>assistant<|end_header_id|>

Olivia had 23 dollars. 5 bagels for 3 dollars each will be 5 x 3 = 15 dollars. So she has 23 - 15 dollars left. 23 - 15 is 8. The final answer is 8<|eot_id|><|start_header_id|>user<|end_header_id|>


\end{lstlisting}

\subsection{MBPP}
\begin{lstlisting}
You are an expert Python programmer, and here is your task:
Write a function to find the similar elements from the given two tuple lists.
Your code should pass the following tests:
assert similar_elements((3, 4, 5, 6),(5, 7, 4, 10)) == (4, 5)
assert similar_elements((1, 2, 3, 4),(5, 4, 3, 7)) == (3, 4)
assert similar_elements((11, 12, 14, 13),(17, 15, 14, 13)) == (13, 14)<|eot_id|><|start_header_id|>assistant<|end_header_id|>

```python
def similar_elements(test_tup1, test_tup2):
    res = tuple(set(test_tup1) & set(test_tup2))
    return (res)
```<|eot_id|><|start_header_id|>user<|end_header_id|>

You are an expert Python programmer, and here is your task:
Write a python function to identify non-prime numbers.
Your code should pass the following tests:
assert is_not_prime(2) == False
assert is_not_prime(10) == True
assert is_not_prime(35) == True<|eot_id|><|start_header_id|>assistant<|end_header_id|>

```python
import math
def is_not_prime(n):
    result = False
    for i in range(2,int(math.sqrt(n)) + 1):
        if n %
            result = True
    return result
```<|eot_id|><|start_header_id|>user<|end_header_id|>

You are an expert Python programmer, and here is your task:
Write a function to find the largest integers from a given list of numbers using heap queue algorithm.
Your code should pass the following tests:
assert heap_queue_largest( [25, 35, 22, 85, 14, 65, 75, 22, 58], 3)==[85, 75, 65]
assert heap_queue_largest( [25, 35, 22, 85, 14, 65, 75, 22, 58], 2)==[85, 75]
assert heap_queue_largest( [25, 35, 22, 85, 14, 65, 75, 22, 58], 5)==[85, 75, 65, 58, 35]<|eot_id|><|start_header_id|>assistant<|end_header_id|>

```python
import heapq as hq
def heap_queue_largest(nums,n):
    largest_nums = hq.nlargest(n, nums)
    return largest_nums
```<|eot_id|><|start_header_id|>user<|end_header_id|>


\end{lstlisting}

\subsection{Arc-Challenge and Arc-Easy}
We use the prompt sourced from self-consistency chain-of-thought \cite{wang2023selfconsistencyimproveschainthought}.

\begin{lstlisting}
Given the following problem, reason and give a final answer to the problem.
Problem: George wants to warm his hands quickly by rubbing them. Which skin surface will produce the most heat?
A. dry palms
B. wet palms
C. palms covered with oil
D. palms covered with lotion
Your response should end with "The final answer is [answer]" where [answer] is the response to the problem.<|eot_id|><|start_header_id|>assistant<|end_header_id|>

Dry surfaces will more likely cause more friction via rubbing than other smoother surfaces, hence dry palms will produce the most heat. The final answer is A.<|eot_id|><|start_header_id|>user<|end_header_id|>

Given the following problem, reason and give a final answer to the problem.
Problem: Which factor will most likely cause a person to develop a fever?
A. a leg muscle relaxing after exercise
B. a bacterial population in the bloodstream
C. several viral particles on the skin
D. carbohydrates being digested in the stomach
Your response should end with "The final answer is [answer]" where [answer] is the response to the problem.<|eot_id|><|start_header_id|>assistant<|end_header_id|>

Option B, bacterial population is the most likely cause for a person developing fever. The final answer is B.<|eot_id|><|start_header_id|>user<|end_header_id|>

Given the following problem, reason and give a final answer to the problem.
Problem: Which change in the state of water particles causes the particles to become arranged in a fixed position?
A. boiling
B. melting
C. freezing
D. evaporating
Your response should end with "The final answer is [answer]" where [answer] is the response to the problem.<|eot_id|><|start_header_id|>assistant<|end_header_id|>

When water is frozen, the particles are arranged in a fixed position; the particles are still moving for all other options. The final answer is C.<|eot_id|><|start_header_id|>user<|end_header_id|>

Given the following problem, reason and give a final answer to the problem.
Problem: When a switch is used in an electrical circuit, the switch can
A. cause the charge to build
B. increase and decrease the voltage
C. cause the current to change direction
D. stop and start the flow of current
Your response should end with "The final answer is [answer]" where [answer] is the response to the problem.<|eot_id|><|start_header_id|>assistant<|end_header_id|>

The function of a switch is to start and stop the flow of a current. The final answer is D.<|eot_id|><|start_header_id|>user<|end_header_id|>


\end{lstlisting}

\subsection{CommonsenseQA}
We use the CSQA prompt sourced from the original chain-of-thought paper \cite{wei2023chainofthought}.

\begin{lstlisting}
Given the following problem, reason and give a final answer to the problem.
Problem: What do people use to absorb extra ink from a fountain pen?
A. shirt pocket
B. calligrapher's hand
C. inkwell
D. desk drawer
E. blotter
Your response should end with "The final answer is [answer]" where [answer] is the response to the problem.<|eot_id|><|start_header_id|>assistant<|end_header_id|>

The answer must be an item that can absorb ink. Of the above choices, only blotters are used to absorb ink. The final answer is E.<|eot_id|><|start_header_id|>user<|end_header_id|>

Given the following problem, reason and give a final answer to the problem.
Problem: What home entertainment equipment requires cable?
A. radio shack
B. substation
C. television
D. cabinet
Your response should end with "The final answer is [answer]" where [answer] is the response to the problem.<|eot_id|><|start_header_id|>assistant<|end_header_id|>

The answer must require cable. Of the above choices, only television requires cable. The final answer is C.<|eot_id|><|start_header_id|>user<|end_header_id|>

Given the following problem, reason and give a final answer to the problem.
Problem: The fox walked from the city into the forest, what was it looking for?
A. pretty flowers
B. hen house
C. natural habitat
D. storybook
Your response should end with "The final answer is [answer]" where [answer] is the response to the problem.<|eot_id|><|start_header_id|>assistant<|end_header_id|>

The answer must be something in the forest. Of the above choices, only natural habitat is in the forest. The final answer is C.<|eot_id|><|start_header_id|>user<|end_header_id|>

Given the following problem, reason and give a final answer to the problem.
Problem: Sammy wanted to go to where the people were. Where might he go?
A. populated areas
B. race track
C. desert
D. apartment
E. roadblock
Your response should end with "The final answer is [answer]" where [answer] is the response to the problem.<|eot_id|><|start_header_id|>assistant<|end_header_id|>

The answer must be a place with a lot of people. Of the above choices, only populated areas have a lot of people. The final answer is A.<|eot_id|><|start_header_id|>user<|end_header_id|>

Given the following problem, reason and give a final answer to the problem.
Problem: Where do you put your grapes just before checking out?
A. mouth
B. grocery cart
C. super market
D. fruit basket
E. fruit market
Your response should end with "The final answer is [answer]" where [answer] is the response to the problem.<|eot_id|><|start_header_id|>assistant<|end_header_id|>

The answer should be the place where grocery items are placed before checking out. Of the above choices, grocery cart makes the most sense for holding grocery items. The final answer is B.<|eot_id|><|start_header_id|>user<|end_header_id|>

Given the following problem, reason and give a final answer to the problem.
Problem: Google Maps and other highway and street GPS services have replaced what?
A. united states
B. mexico
C. countryside
D. atlas
Your response should end with "The final answer is [answer]" where [answer] is the response to the problem.<|eot_id|><|start_header_id|>assistant<|end_header_id|>

The answer must be something that used to do what Google Maps and GPS services do, which is to give directions. Of the above choices, only atlases are used to give directions. The final answer is D.<|eot_id|><|start_header_id|>user<|end_header_id|>

Given the following problem, reason and give a final answer to the problem.
Problem: Before getting a divorce, what did the wife feel who was doing all the work?
A. harder
B. anguish
C. bitterness
D. tears
E. sadness
Your response should end with "The final answer is [answer]" where [answer] is the response to the problem.<|eot_id|><|start_header_id|>assistant<|end_header_id|>

The answer should be the feeling of someone getting divorced who was doing all the work. Of the above choices, the closest feeling is bitterness. The final answer is C.<|eot_id|><|start_header_id|>user<|end_header_id|>


\end{lstlisting}

\section{Question dataset about Pavel Durov's arrest and release}
\label{sec:pavel_questions}
We handcrafted a dataset of 20 questions to evaluate our model after prompting vs. after baking in news headlines relating to Pavel Durov. We present the full question set here:

\begin{enumerate}
    \item Is Pavel Durov in custody in France as of August 28th, 2024?
    \item What did the New York Times report about Pavel Durov on August 28th, 2024?
    \item Did the Telegram founder ever get arrested in France?
    \item Did French authorities release Pavel Durov from custody?
    \item On what day was Pavel Durov released from custody?
    \item Has the Telegram founder been in legal trouble in the month of August, 2024?
    \item Where was Pavel Durov arrested in August of the year 2024?
    \item What did France set Pavel Durov's bail to?
    \item Did Pavel Durov post bail on August 29th, 2024?
    \item Who paid a 5 million dollar bail on August 29th, 2024?
    \item What is the latest on Pavel Durov?
    \item Who was charged with a wide range of crimes in France on August 28th, 2024?
    \item What is the probe into Telegram and Pavel Durov in August 2024?
    \item What founder of a popular messaging app was arrested and charged with a range of crimes in France in August 2024?
    \item What happened to Telegram's founder in France in mid 2024?
    \item Was Elon Musk arrested in France on August 28th, 2024?
    \item Was the founder of Twitter arrested by authorities?
    \item Was Pavel Durov arrested and charged with crimes in Serbia?
    \item Do French authorities, as of August 28th, believe Pavel Durov innocent?
    \item Did Jack Dorsey get arrested by French authorities?
\end{enumerate}

\end{document}